\documentclass[runningheads]{llncs}
\usepackage{graphicx}
\usepackage{amsmath,amssymb} 
\usepackage{color}
\usepackage[width=122mm,left=12mm,paperwidth=146mm,height=193mm,top=12mm,paperheight=217mm]{geometry}

\usepackage[utf8]{inputenc}

\usepackage{tikz}
\usepackage{comment}

\usepackage{enumerate}
\usepackage{bm}
\usepackage[ruled,linesnumbered,vlined]{algorithm2e}
\usepackage{multirow}
\usepackage{tabularx} 
\usepackage{booktabs}
\usepackage[font=small]{caption}
\usepackage{adjustbox}
\usepackage{subcaption}

\DeclareMathOperator*{\argmin}{arg\,min}

\newcommand{\bt}{\mathbf{t}}

\newcommand{\bR}{\mathbf{R}}

\newcommand{\plucker}{Pl$\ddot{\text{u}}$cker}
\newcommand{\eg}{\emph{e.g.}}
\newcommand{\ie}{i.\,e.}
\definecolor{ForestGreen}{RGB}{34,139,34}

\newcommand{\figref}[1]{Figure~\ref{#1}}

\SetKwInput{KwData}{Inputs}
\SetKwInput{KwResult}{Output}
\SetKwComment{Comment}{$\triangleright$\ }{}
\DeclareMathAlphabet\mathbfcal{OMS}{cmsy}{b}{n}
\usepackage{blindtext}
\usepackage{multicol}
\usepackage{wrapfig}

\begin{document}
\pagestyle{headings}
\mainmatter
\def\ECCV16SubNumber{***}  

\title{CyberLoc: Towards Accurate Long-term Visual Localization} 

\titlerunning{CyberLoc}

\authorrunning{ }

\author{\footnotesize Liu Liu \thanks{Equal contributions}, Yukai Lin $^\star$, Xiao Liang $^\star$, Qichao Xu $^\star$, Miao Jia, Yangdong Liu, Yuxiang Wen, Wei Luo, Jiangwei Li }
\institute{Cyberverse Dept, Huawei Cloud Computing Technologies Co., Ltd.}

\maketitle

\begin{abstract}
This technical report introduces \textit{CyberLoc}, an image-based visual localization pipeline for robust and accurate long-term pose estimation under challenging conditions. 
The proposed method comprises four modules connected in a sequence. First, a mapping module is applied to build accurate 3D maps of the scene, one map for each reference sequence if there exist multiple reference sequences under different conditions. Second, a single-image-based localization pipeline (retrieval--matching--PnP) is performed to estimate 6-DoF camera poses for each query image, one for each 3D map. Third, a consensus set maximization module is proposed to filter out outlier 6-DoF camera poses, and outputs one 6-DoF camera pose for a query. Finally, a robust pose refinement module is proposed to optimize 6-DoF query poses, taking candidate global 6-DoF camera poses and their corresponding global 2D-3D matches, sparse 2D-2D feature matches between consecutive query  images and SLAM poses of the query sequence as input.
Experiments on the 4seasons dataset show that our method achieves high accuracy and robustness.
In particular, our approach wins the localization challenge of ECCV 2022 workshop on Map-based Localization for Autonomous Driving (MLAD-ECCV2022). 

\end{abstract}

\begin{figure}
\centering
\includegraphics[width=0.9\textwidth]{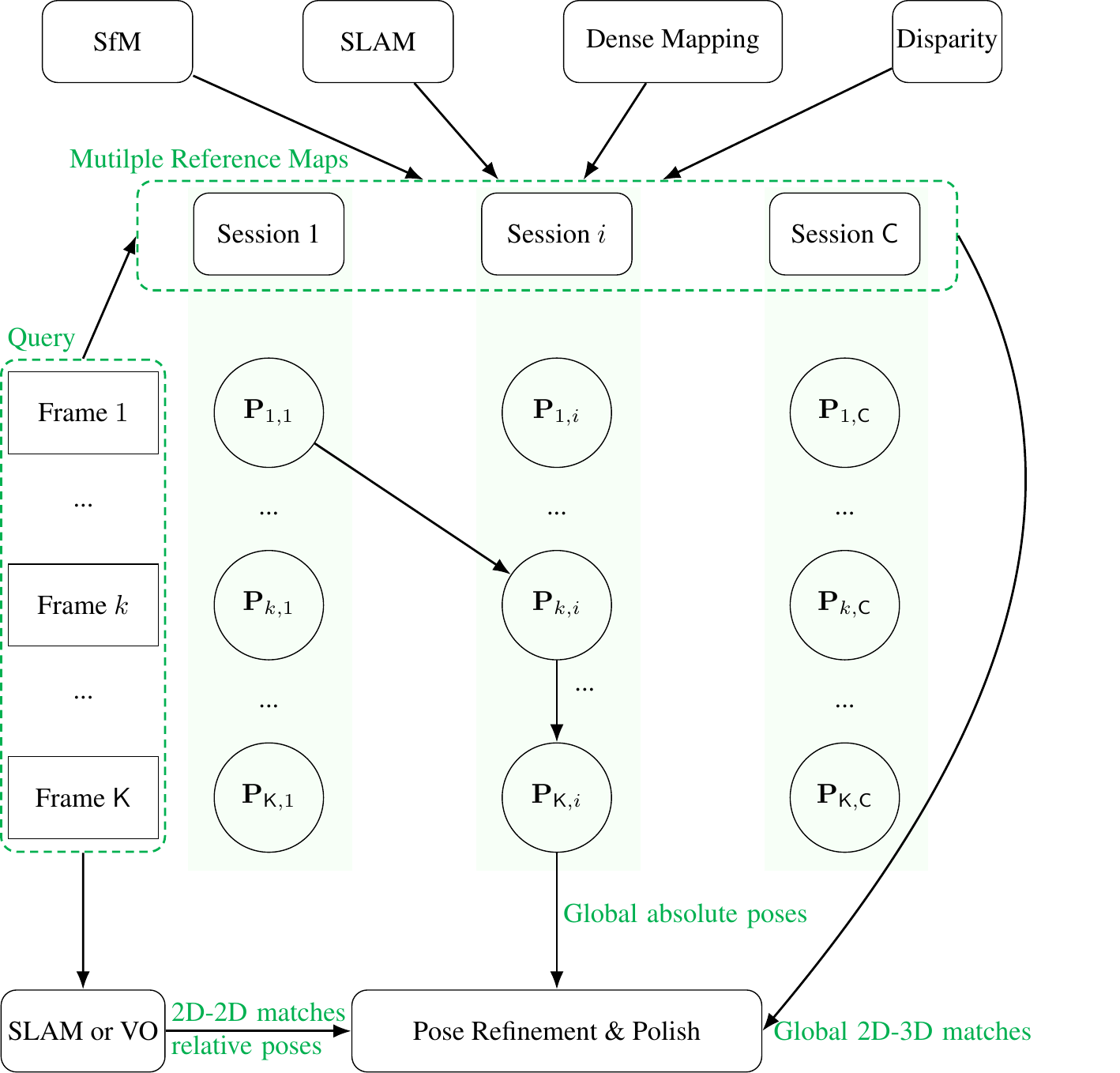}
\caption{\it The overall framework of our method. A scene can be visited multiple times, resulting in multiple reference maps under different conditions (weather, lighting, etc.). For each reference sequence, we perform stereo SfM, SLAM, dense mapping, and disparity to reconstruct four 3D maps for robustness. Given a query frame, we first perform localization with respect to each session map and obtain multiple global poses corresponding to sessions (one global absolute pose per session). We then use a consensus set maximization module to select the best poses for query frames (one pose per query). We further use a SLAM module to obtain 2D-2D matches and relative poses between consecutive query frames. Finally, with 2D-2D matches, relative pose, global absolute poses and their corresponding global 2D-3D matches, we use a pose refinement module to optimize the query poses. A pose polish module can be optionally used for better performance.     }
\label{fig::overall_framework}
\end{figure}

\section{Introduction}

This technical report studies the problem of image-based localization with respect to a pre-built 3D map.  This problem has attracted considerable attention recently due to the widespread potential applications, such as in autonomous driving \cite{wenzel20204seasons}, robotics~\cite{strisciuglio2018trimbot2020} and VR/AR \cite{sarlin2022lamar}. It aims to estimate the 6-DoF global pose for a query image given a pre-built 3D map. Although visual localization has progressed rapidly in the past few years, how to achieve a robust and accurate localization under long-term challenging conditions still remains to be solved.

The main challenges are: 1) how to create accurate 3D maps that are robust to environmental changes, and 2) how to use the pre-built maps to accurately localize the camera.

To address the first challenge, we propose to use a stereo-camera rig with GPS-IMU to mapping the world. Given stereo image sequences with ground-truth 6-DoF poses, we separately reconstruct four 3D maps using four methods: 1) stereo Structure from Motion (SfM) with sparse 2D image features; 2) stereo SLAM; 3) stereo SfM with dense image matching; and 4) stereo disparity for each frame. Our motivation of using the above four 3D methods is to build a robust 3D map with respect to scene changes.

To address the second challenge, we introduce two modules, namely the consensus set maximization and pose refinement module. Given multiple global poses corresponding to multiple reference maps, the consensus set maximization aims to select the best pose for each query, resulting in one global pose and one set of global 2D-3D matches per query image. For query image sequences, the pose refinement module utilizes the global information (\ie, the global poses and 2D-3D matches), and the local information (\ie, SLAM poses and 2D-2D matches between consecutive query images), to further optimize query poses. The entire localization pipeline of \textit{CyberLoc} is given in Figure~\ref{fig::overall_framework}.

The \textbf{main contributions} of this technical report  are:
\begin{enumerate}
    \item We propose a visual localization pipeline consisting of four consecutive modules that help to achieve high accuracy and robustness under long-term environmental changes;
    \item We show that using multiple reference maps helps to overcome failed localization caused by long-term scene changes. This is achieved by a new consensus set maximization module that identifies the best query pose with respect to multiple reference maps;
    \item We introduce a robust pose refinement method, combining global information from pre-built maps and local information from SLAM, to refine query poses.
\end{enumerate}
The proposed method is validated on the 4seasons dataset \cite{wenzel20204seasons} and achieves state-of-the-art performance.
In the following sections, we will give details of the proposed four modules. We first give our mapping pipeline in Sec.~\ref{sec::mapping}, to reconstruct 3D maps for multiple reference sessions. Next, we present our single image localization in Sec.~\ref{sec::Localization}, to obtain multiple global poses for each query image. We then give the proposed consensus set maximization method in Sec.~\ref{sec::Multi-sessionCSM}, to select the best query pose for each query image. Finally, in Sec.~\ref{sec::pose_refinement}, we provide the proposed robust pose refinement method.


\section{Mapping}\label{sec::mapping}

\subsection{Image Pre-processing}

In this section, we introduce some image pre-processing steps before using images for sparse 3D map reconstruction.

\begin{figure}
		\begin{subfigure}{.5\textwidth}
			\centering
			\includegraphics[width=\textwidth]{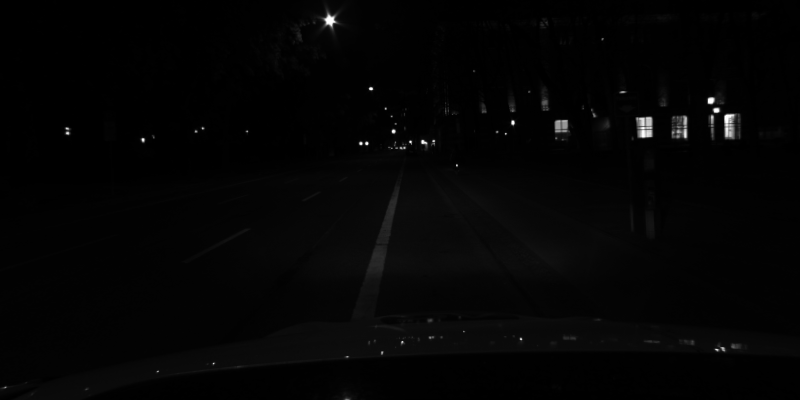}
			\caption{low light image}
		\end{subfigure}
		\begin{subfigure}{.5\textwidth}
			\centering
			\includegraphics[width=\textwidth]{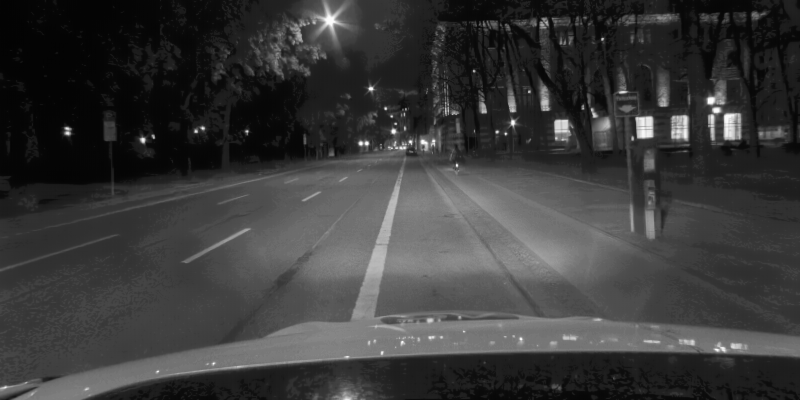}
			\caption{enhanced image}
		\end{subfigure}
		\caption{\it An example of low light image enhancement.}
		\label{fig:LLFlow}
\end{figure}

\paragraph{Low light image enhancement.}
For images captured in the night, we perform low light image enhancement using LLFlow \cite{wang2021low}. An example is given in Figure \ref{fig:LLFlow}. We find that using enhanced images would help to extract better 2D local  features and global feature vectors from images. 
\paragraph{Semantic segmentation.}
When working in highly dynamic environment, both SfM and SLAM methods would perform poorly due to interference from dynamic objects \cite{liu2017robust}. Feature points on  dynamic objects are either unrepeatable when performing 3D reconstruction or `fool' the feature tracking in SLAM. Following common practice \cite{saputra2018visual}, we perform semantic segmentation using SegFormer \cite{xie2021segformer} to mask out dynamic objects (\eg, car, cyclist, etc.). Furthermore, pixels belong to the car shield are also masked out.

\paragraph{Resizing.}
Before using images to extract features, we resize images while keeping their original aspect ratio. The reason is that the performance of feature extraction networks is vulnerable to image size. To extract a global feature vector from an image, we use a size of $800$ (long side). To extract 2D local features from an image, we use a size of $2000$. The two values are used as they perform best on the validation dataset.

\subsection{Feature Extraction}\label{sec::Feature_Extraction}

\paragraph{Global feature vector.} Global feature vectors of images are used by image retrieval to, 1) find co-visible images for an image in the mapping (\ie, SfM) stage; or 2) find a local map for each query image in the localization stage. To extract discriminative global feature vectors, we use an internal neural network trained on open street-view images and internal datasets. The comparison with respect to state-of-the-art NetVLAD \cite{arandjelovic2016netvlad}, SARE \cite{liu2019stochastic}, SFRS \cite{ge2020self} on typical benchmarks and the 4seasons dataset is given in Table \ref{tbl:retrieval}. 

\begin{table}[]
\caption{\it Comparison of our image retrieval method with respect to state-of-the-art methods. We use the same distance threshold (25m) and evaluation metric as \cite{arandjelovic2016netvlad} to determine whether an image is successfully localized. Our model consistently outperform other methods on standard benchmarks and the 4seasons dataset.}
\vspace*{2mm}
\label{tbl:retrieval}
\centering
\resizebox{0.95\textwidth}{!}{
\begin{tabularx}{435pt}{c *{12}{>{\centering\arraybackslash}X}}
\hline
{Methods} & \multicolumn{2}{c}{{Pitts250k}} & \multicolumn{2}{c}{{Tokyo 24/7}} & \multicolumn{2}{c}{{St Lucia}} & \multicolumn{2}{c}{{Cityloop}} & \multicolumn{2}{c}{{Oldtown}} & \multicolumn{2}{c}{{Parkinggarage}} \\
& R@1                & R@5               & R@1                & R@5                & R@1               & R@5               & R@1               & R@5               & R@1               & R@5              & R@1                  & R@5 \\
\hline
NetVLAD\cite{arandjelovic2016netvlad} & 86.0 & 93.2 & 73.3 & 82.9  & -  & -  & -  & -  & -    & -  & -  & - \\
SARE\cite{liu2019stochastic} & 89.0 & 95.5 & 79.7 & 86.7 & - & - & - & - & - & - & - & -  \\
SFRS\cite{ge2020self} & 90.7 & 96.4 & 85.4 & 91.1 & 86.1 & 93.5 & 73.3 & 82.4 & 32.8 & 44.8 & 97.2 & 99.3 \\
\textbf{Ours} & \textbf{93.0} & \textbf{97.5} & \textbf{89.5} & \textbf{94.6} & \textbf{99.6} & \textbf{99.9} & \textbf{95.9} & \textbf{97.2} & \textbf{67.7} & \textbf{74.7} & \textbf{99.7}  & \textbf{100.} \\ \hline
\end{tabularx}
}
\end{table}

Assembling multiple Global feature vectors from multiple networks is a common practice to improve the performance of image retrieval \footnote{Please refer to the Google landmark retrieval challenge for more information}. However, we found that the assembling technique does not work on the 4seasons dataset. The reason is that our single model outperforms other state-of-the-art methods by a large margin on the 4seasons dataset. 

\paragraph{Local features.} 2D local features are used to perform sparse 3D reconstruction of the environment. We extract around 2000 SuperPoint \cite{detone2018superpoint} feature points for an image.

\subsection{SfM} \label{sec::Stereo_Sparse_Triangulation}
We have pre-recorded database images with ground-truth 6-DoF camera poses in a world coordinate system. We perform image retrieval to find Top-K (K=$40$) similar database images. For each database image pair, we perform sparse feature matching and find 2D-2D matches. We further prune out wrong 2D-2D matches by enforcing the epipolar constraint using the ground-truth poses. With 2D-2D matches and ground-truth poses, we triangulate \footnote{ SVD with post non-linear refinement.} 3D points and perform structure-only bundle adjustment \footnote{Fixing camera poses.} to refine the positions of 3D points using colmap \cite{schonberger2016structure}.


\paragraph{Remark 1.} Given Superpoint feature points, though SGMNet \cite{chen2021learning}, ClusterGNN \cite{shi2022clustergnn} and ELA \cite{suwanwimolkul2022efficient} successfully reduce the computation complexity, their performance is worse than the pre-trained SuperGlue \footnote{See results on the YFCC100M \cite{thomee2016yfcc100m} and FM-Bench \cite{bian2019evaluation} datasets.}. Recently, our proposed FGCNet \cite{FGCNet} achieves 4x speedup than SuperGlue, while maintaining competitive performance with respect to the pre-trained SuperGlue. 

\paragraph{Remark 2.} For a triangulated 3D point, its precision depends on point-camera distances, number of visible cameras, view angles, etc. Though MegLoc \cite{peng2021megloc} proposes to prune out 3D points with large uncertainty \footnote{Please refer to Sec.5.1 of \cite{zou2012coslam}.}, we found the uncertainty threshold is hard to tune and we decided not to use this filtering step.

\subsection{SLAM}
Given ground-truth 6-DoF camera poses of images,
with SuperPoint feature points, we perform SLAM using the ptam \cite{pire2017s}. Note that we skip the pose estimation stage in the SLAM and only optimize 3D points in the local map optimization stage.

\paragraph{Remark.} The main difference between 3D points from SfM and SLAM is that we use ALL images in SLAM rather than keyframes in SfM. 3D points are triangulated and updated sequentially,  rather than in a bundle.

\subsection{Dense Mapping}
Using the same retrieved pairs in Sec.\ref{sec::Stereo_Sparse_Triangulation}, we use dense feature matching method QTA \cite{tang2022quadtree} to build sparse 2D-2D matches and then triangulate 3D points in the same way as Sec.\ref{sec::Stereo_Sparse_Triangulation}. Since dense feature matching methods do not detect sparse 2D feature points and can generate unrepeatable 2D-2D matches \footnote{Given 2D-2D matches for image pair A-B and A-C, 2D feature points in image A are different.}, we use the same quantization technique in Patch2Pix \cite{zhou2021patch2pix} to alleviate this problem.

\paragraph{Remark.} The main difference between 3D points from Dense Mapping and SfM is that we use dense feature matching method QTA \cite{tang2022quadtree} rather than sparse feature matching method SuperGlue \cite{sarlin2020superglue} to build 2D-2D matches. The motivation is that QTA \cite{tang2022quadtree} can generate more matches than SuperGlue \cite{sarlin2020superglue}  for textureless image pairs. More 3D points can be triangulated, resulting in a robust 3D representation for textureless areas.

\subsection{Disparity} 
\begin{figure}
\centering
\includegraphics[width=0.6\textwidth]{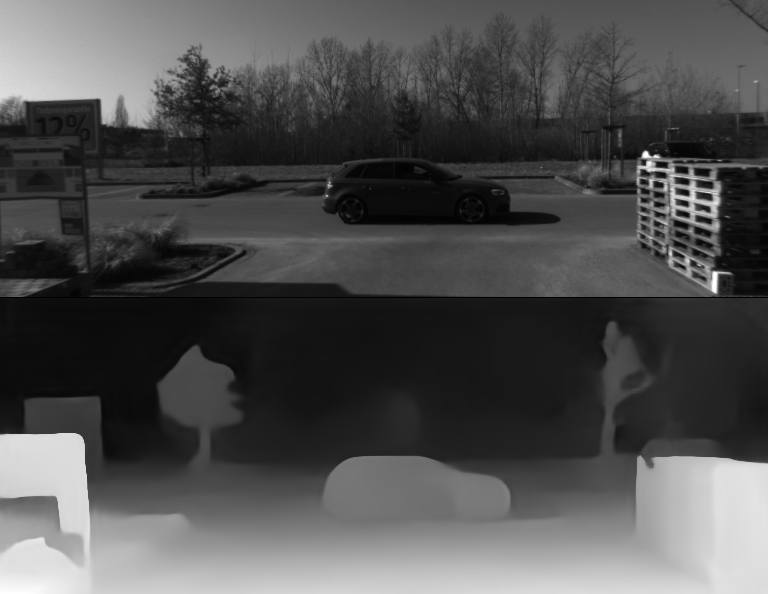}
\caption{\it An example of disparity estimation. Top: a database image; Bottom: the corresponding disparity image.}
\label{fig::disparity}
\end{figure}

For each image pair, we can estimate a dense disparity map using the pre-trained LEAStereo \cite{cheng2020hierarchical}, thus obtaining a local 3D map for each database image. An example of the estimated  disparity map is given in \figref{fig::disparity}.

\section{Localization}\label{sec::Localization}
For each query image, we retrieve a set of candidate database keyframes using the same global feature vectors described in Sec.\ref{sec::Feature_Extraction}. Since we have four pre-reconstructed 3D maps from SfM, SLAM, dense mapping, and disparity, we first perform four independent localization steps and then combine these localization results.

Specifically, for maps from SfM and SLAM, we first find 2D-2D matches using SuperGlue for each query and candidate database pair. Since we have recorded the matching 3D point (if have) of each database 2D feature point, a set of 2D-3D matches are automatically obtained for each query and candidate database image pair.  

For maps from dense mapping, the only difference is that we use the dense feature matching method QTA \cite{tang2022quadtree} to find 2D-3D matches and all other components are the same.

For maps from disparity, we also use QTA \cite{tang2022quadtree}  to find 2D-2D matches. For each 2D feature point from the database image, we triangulate 3D points using the disparity map. With 2D-2D matches and 3D points for database feature points, we obtain a set of 2D-3D matches. 

Finally, the 2D-3D matches from the top-K ($K=40$) candidate database images and above four maps are combined to perform PnP \footnote{Generalized camera model, aka, \plucker\ lines \cite{pless2003using}. 
} to estimate a 6-DoF camera pose in the world coordinate system.

\paragraph{Remark 1.} The motivation of using 2D-3D matches from different maps is to utilize their complimentary characteristics to improve the robustness of our method with respect to large scene changes.

\paragraph{Remark 2.} Directly combining all 2D-3D matches from top-K ($K=40$) candidate database keyframes and above four maps would generate a large set of 2D-3D matches with low inlier ratio, posing significant challenge for the subsequent RANSAC-PnP step. To improve the inlier ratio, for each map, we perform RANSAC-PnP to obtain a set of inlier 2D-3D matches, and remove duplicated 2D-3D matches \footnote{One 2D feature point from the query image may be matched to multiple different 3D points. We only keep one 2D-3D pair with minimal reprojection error.} before combing 2D-3D matches from four maps.

\section{Multi-session Consensus Set Maximization}\label{sec::Multi-sessionCSM}
In a scene with multiple reference sequences such as the 4seasons\cite{wenzel20204seasons} dataset, multi-session maps can be generated for the scene. In this section, we introduce a consensus set maximization method to fuse localization results using these  multi-session maps.

A simple method to use multi-session maps is to combine 2D-3D matches from multi-session maps to perform RANSAC-PnP. Although it works for complementary multi-session results, the performance is limited since the best localization result from one-session map can be worsened by 2D-3D matches from other-session maps. Using combined 2D-3D matches for RANSAC-PnP would prone to produce averaged (or one dominant)  6-DoF pose, for poses from multi-session maps.

Another method to use multi-session maps is to find the best combination of multi-session maps using trial-and-error \cite{labbe2021multi}. It works for multi-session
maps captured in different times of a day, in small-scale room environment. For the 4seasons dataset, the large-scale multi-session maps are captured in different times of a year, making it hard to find the best combination.

The key idea of our method is finding the best 6-DoF pose from multi-session maps for each query image through an optimization process, rather than combining 2D-3D matches from multi-session maps to perform RANSAC-PnP.
The proposed method consistently produces optimal results across different datasets.

\subsection{Problem Definition}
Let $\mathbf{I}_\mathsf{K} = \{\mathbf{I}_k |  k = 1,\cdots,\mathsf{K}\}$ denotes a query image sequence up to timestamp $\mathsf{K}$. For each query image $\mathbf{I}_k$, we have $\mathsf{C}$ candidate 6-DoF poses $\mathcal{P}_k = \{\mathbf{P}_{k,i} = (\bR_{k,i}, \bt_{k,i}) |  i = 1,\cdots,\mathsf{C}\}$, one from a reference map. 

We aim to find the most accurate 6-DoF pose $\mathbf{P}_{k,i^{*}}$ for $\mathbf{I}_k$.

\subsection{Consensus Set Maximization}
For query images $\mathbf{I}_m$ and $\mathbf{I}_n$, we can obtain a set of 2D-2D matches $\mathcal{Q}_{m,n}$ via feature tracking in SLAM. 
Given query poses $\mathbf{P}_{m,i}$ and $\mathbf{P}_{n,j}$, we can compute the number of inlier 2D-2D matches subject to $\mathbf{P}_{m,i}$ and $\mathbf{P}_{n,j}$, by thresholding the Sampson errors \cite{hartley2003multiple} of 2D-2D matches $\mathcal{Q}_{m,n}$. 
We denote the number of inlier 2D-2D matches as $S_{m,i,n,j}$ and use it to measure the compatibleness of poses $\mathbf{P}_{m,i}$ and $\mathbf{P}_{n,j}$. 
We assume the best poses would generate  the largest  score $\sum S_{m,i,n,j}$, \ie, finding the largest consensus set. Though the problem is non-convex \cite{li2009consensus} and the original $\mathcal{Q}_{m,n}$ contains outlier matches, we found the score $\sum S_{m,i,n,j}$ describes the correctness of poses well.
The consensus set maximization problem is solved by a integer programming process.

\subsection{ Integer Programming}

\begin{figure}[tbp]
\centering
\includegraphics[width=0.6\textwidth]{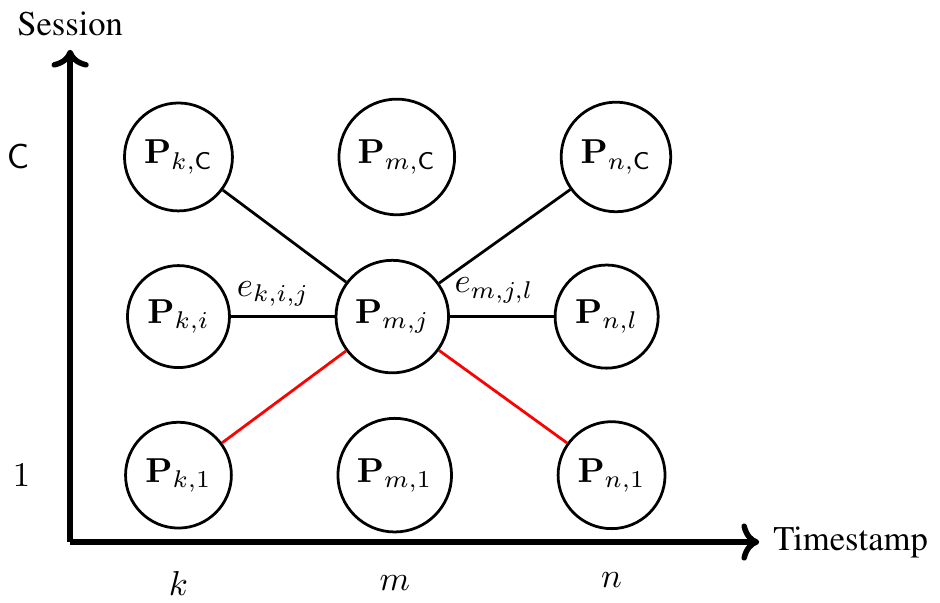}
\caption{\it We use three frames $k,m,n$  to describe our integer programming process. Given poses $\mathcal{P}_k$, $\mathcal{P}_m$, $\mathcal{P}_n$ from multi-session maps at timestamps $k,m,n$, respectively, we aim to find the best poses connected by the red edges. 
Edges $e_{k,i,j} \in \{0,1\}$ denotes the connectivity between nodes (poses) $\mathbf{P}_{k,i}$ and $\mathbf{P}_{m,j}$. $e_{k,i,j}$ equals $1$ only when both nodes $\mathbf{P}_{k,i}$ and $\mathbf{P}_{m,j}$ are selected (deemed as the best poses). 
For the sake of clarity, we only draw edges connecting the node $\mathbf{P}_{m,j}$. Note that all nodes are fully connected for consecutive timestamps. }
\label{fig::inter_prog}
\end{figure}

We use three frames $k,m,n$  to describe our integer programming process, as is given in Figure \ref{fig::inter_prog}. We first build a densely connected graph for consecutive timestamps. For nodes $\mathbf{P}_{k,i}$ and $\mathbf{P}_{m,j}$, there is an edge $e_{k,i,j} \in \{0,1\}$ connecting them. The score of edge $e_{k,i,j}$ is $S_{k,i,m,j}$ (abbrev. $S_{k,i,j}$ for clarity), which is the number of inlier 2D-2D matches using poses $\mathbf{P}_{k,i}$ and $\mathbf{P}_{m,j}$. Our integer programming process is given by,
\begin{align}
 {}& \argmin_{\mathcal{E}=\{e_{k,i,j}\}}\sum_{k=1}^{\mathsf{K}-1}\sum_{i=1}^{\mathsf{C}}\sum_{j=1}^{\mathsf{C}}-e_{k,i,j}S_{k,i,j}, \\
  s.t. & \sum_{i=1}^{C}\sum_{j=1}^{C}e_{k,i,j} = 1, \label{eq::constraint1} \\
  & \sum_{i=1}^{C}e_{k,i,j} = \sum_{l=1}^{C}e_{m,j,l}.\label{eq::constraint2}
\end{align}

Eq.\eqref{eq::constraint1} enforces that exactly one edge should be selected for the connectivity of the graph.   Eq.\eqref{eq::constraint2} enforces the following relationship: 1) if node $\mathbf{P}_{m,j}$ is selected, there must be an edge connecting $\mathbf{P}_{m,j}$ to other nodes in consecutive frames; 2) if node $\mathbf{P}_{m,j}$ is NOT selected, all edges connecting $\mathbf{P}_{m,j}$ to other nodes in consecutive frames should be removed.

There are a maximum \footnote{If one map session fails to estimate a camera pose, the number of edges between consecutive timestamps is smaller than $\mathsf{C} \times \mathsf{C}$.} of $(\mathsf{K}-1)\times \mathsf{C} \times \mathsf{C}$ edges (optimization variables) in our integer programming. The solution can be found very efficiently by off-the-shelf toolboxes. After optimization, nodes (poses) connected by edges  $e_{k,i,j} = 1$ are deemed as the best poses.

\subsection{Experimental Result}
\label{sec:cons_res}

We validate the proposed consensus set maximization method on the oldtown and cityloop datasets, from the 4seasons dataset.  
The number of image sequences with ground-truth poses is four and three, for the oldtown and  cityloop dataset, respectively. For each dataset, one sequence is used for testing and the rest sequences are used for mapping.
The localization results are separately given in Table \ref{tbl:consensus1} and Table \ref{tbl:consensus2}. The results show that the proposed consensus set maximization method is robust and consistently shows good performance on the two datasets. In contrast, the simple method of merging 2D-3D matches does not work on the cityloop dataset. The reason is that localization results using multi-session maps are not compatible, on the cityloop dataset. 
\begin{table}[tbp]
    \caption{\it Result of the proposed consensus set maximization on the oldtown dataset. We report localization recalls with respect to different translation error thresholds. 
    Both merging 2D-3D matches from multi-session maps and our integer programming method achieve the best performance.}
    \vspace*{2mm}
    \label{tbl:consensus1}
    \centering
    \resizebox{0.95\textwidth}{!}{%
    \begin{tabularx}{\textwidth}{c *{5}{>{\centering\arraybackslash}X}} 
{Trans.Err.} & {Ref\_0} & {Ref\_1} & {Ref\_2} & {2D-3D Merge} & {Int. Prog.} \\ \hline
0.05m               & 36.60\%         & 39.80\%         & 43.20\%         & \textbf{51.00\%}     & 49.80\%             \\
0.1m                & 60.80\%         & 68.40\%         & 69.20\%         & 78.30\%              & \textbf{78.40\%}    \\
0.2m                & 79.30\%         & 87.00\%         & 85.60\%         & \textbf{94.10\%}     & 93.20\%             \\
0.5m                & 89.20\%         & 93.50\%         & 92.30\%         & \textbf{96.60\%}     & 96.50\%             \\
1.0m                & 92.00\%         & 95.50\%         & 94.10\%         & \textbf{97.80\%}     & \textbf{97.80\%}    \\
3.0m                & 93.70\%         & 96.50\%         & 95.00\%         & \textbf{98.30\%}     & \textbf{98.30\%}    \\ \hline
    \end{tabularx}
    }

\end{table}
\begin{table}[tbp]
    \caption{\it Result of the proposed consensus set maximization on the cityloop dataset. Our integer programming method achieves the best performance. In contrast, simply merging 2D-3D matches from multi-session maps does not work. Its performance is even worse than single-session localization.}
    \vspace*{2mm}
    \label{tbl:consensus2}
    \centering
    \resizebox{0.9\textwidth}{!}{%
    \begin{tabularx}{\textwidth}{c *{4}{>{\centering\arraybackslash}X}} 
{Trans.   Err.} & {Ref\_0} & {Ref\_1} & {2D-3D Merge} & {Int. Prog.} \\ \hline
0.05m                  & 69.20\%         & 74.20\%         & 73.30\%              & \textbf{79.20\%}    \\
0.1m                   & 91.80\%         & 88.00\%         & 87.30\%              & \textbf{92.30\%}    \\
0.2m                   & 98.00\%         & 94.70\%         & 94.20\%              & \textbf{97.70\%}    \\
0.5m                   & 99.60\%         & 99.20\%         & 99.10\%              & \textbf{99.70\%}    \\
1.0m                   & 99.90\%         & 99.50\%         & 99.60\%              & \textbf{100.00\%}   \\
3.0m                   & \textbf{100.00\%}        & 99.70\%         & 99.70\%              & \textbf{100.00\%}   \\ \hline
\end{tabularx}%
    }

\end{table}



We run our method on a PC equipped with a 2TB RAM and an AMD EPYC 7742 CPU. The running time of our method is given in Table~\ref{tab:con_runtime}. The implementation is based on Python and uses CBC solver for optimization. The running time can be significantly reduced using C++.

\begin{table}[tbp]
\centering
\caption{Running time of the proposed consensus set maximization. For the oldtown dataset, we have three reference sequences ($C=3$), and each of the sequence has 3296 frames ($K=3296$). For the cityloop dataset, $C=2$ and $K=10224$.}
\label{tab:con_runtime}
    \vspace*{2mm}

\resizebox{0.7\textwidth}{!}{%
\begin{tabularx}{0.8\textwidth}{ll *{2}{>{\centering\arraybackslash}X}} 
{Dataset}  && {Total time} & {Single frame time} \\ \toprule
Oldtown &&  24.4s &  7.4ms \\ \hline
Cityloop  &&  70.7s &   6.9ms \\ \bottomrule
\end{tabularx}
}
\end{table}

\section{Pose Refinement}\label{sec::pose_refinement}
In this section, we show how to combine global information from Sec. \ref{sec::Multi-sessionCSM} and local information from SLAM to refine query poses.

\subsection{Problem Definition}

For each query image $\mathbf{I}_k$, we have obtained its best global pose $\mathbf{P}_k$ from the multi-session localization step (Sec.~\ref{sec::Multi-sessionCSM}). Considering that the accuracy (even being an outlier) of $\mathbf{P}_k$ varies  for different frames, we associate a weight variable $w_k \in [0,1]$ for each  $\mathbf{P}_k$, describing the weight of $\mathbf{P}_k$ in our optimization process. We also retrieve 2D-3D matches $\mathcal{F}_k$ corresponding to $\mathbf{P}_k$, and denote the reprojection error at pose $\mathbf{X}_k$ as $\pi\left ( \mathbf{X}_k, \mathcal{F}_k\right )$.

For consecutive frames, we can obtain their local relative poses $\mathbf{Z}_{k-1,k}$ and 2D-2D matches $\mathcal{Q}_{k-1,k}$ through SLAM. Given query poses $\mathbf{X}_{k-1}$ and $\mathbf{X}_k$, we denote the two-view matching (Sampson) error as $\rho \left ( \mathbf{X}_{k-1}, \mathbf{X}_k, \mathcal{Q}_{k-1,k} \right )$.

\begin{figure}[tbp]
\centering
\includegraphics[width=0.7\textwidth]{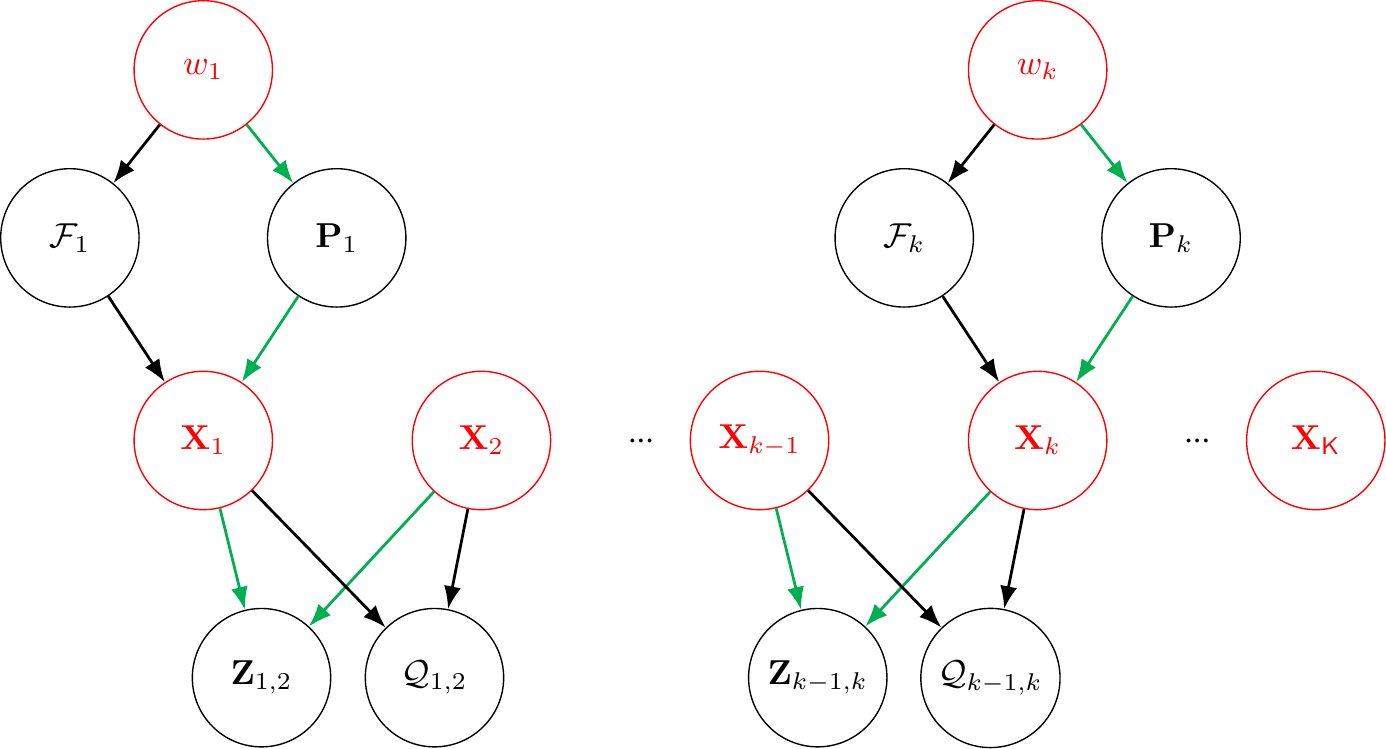}
\caption{\it The framework of our pose refinement step. \{$\mathbf{X}_1$,  ..., $\mathbf{X}_k$, $\mathbf{X}_\mathsf{C}$\} denotes query poses. $\mathbf{P}_k$ denotes the best pose from multi-session localization (Sec.~\ref{sec::Multi-sessionCSM}). $\mathcal{F}_k$ denotes the 2D-3D matches corresponding to the best pose $\mathbf{P}_k$.  $w_k \in [0,1]$  separately denotes the weight of $\mathbf{P}_k$. $\mathbf{Z}_{k-1,k}$ denotes the relative pose from SLAM.  $\mathcal{Q}_{k-1,k}$ denotes the 2D-2D matches from SLAM. Green lines denote a pose-graph in our framework. We aim to solve for red variables.  }
\label{fig::pgba}
\end{figure}

Given query images $\mathbf{I}_\mathsf{K} = \{\mathbf{I}_k |  k = 1,\cdots,\mathsf{K}\}$ up to timestamp $\mathsf{K}$, we aim to solve query poses $\mathcal{X}_\mathsf{K} = \{\mathbf{X}_k |  k = 1,\cdots,\mathsf{K}\}$ and latent weights $\mathcal{W}_\mathsf{K} = \{w_k |  k = 1,\cdots,\mathsf{K}\}$. The overall framework of our optimization problem is given in Fig.~\ref{fig::pgba}.

Our objective function is given by,
\begin{align}\label{eq::loss}
    {}\argmin_{\mathcal{X}_\mathsf{K}, \mathcal{W}_\mathsf{K}}\sum_{k}&w_k\left (\left \| \mathbf{P}_k - \mathbf{X}_k \right \|^2 + \lambda_1 \pi\left ( \mathbf{X}_k, \mathcal{F}_k\right )  \right ) + \notag\\ 
    &\left \|h\left (\mathbf{X}_{k-1}, \mathbf{X}_k  \right ) - \mathbf{Z}_{k-1,k}  \right \|^2  + \lambda_2\rho \left ( \mathbf{X}_{k-1}, \mathbf{X}_k, \mathcal{Q}_{k-1,k} \right ),
\end{align}
where $\left \| \cdot \right \|^2$ denotes the pose error, and $h\left (\mathbf{X}_{k-1}, \mathbf{X}_k  \right )$ denotes the relative pose between $\mathbf{X}_{k-1}$ and $ \mathbf{X}_k$, $\lambda_1$ and $\lambda_2$ are two weights to balance the 2D-3D reprojection error and 2D-2D Sampson error.

\paragraph{Remark 1} If no global pose can be obtained at frame $k$, one can simply remove $\mathcal{F}_k$,  $\mathbf{P}_k$, and $w_k$.
\paragraph{Remark 2} One can also add skip constrains $\mathbf{Z}_{m,n}$ and $\mathcal{Q}_{m,n}$ ($n \neq  m+1$).

\subsection{Robust Optimization}

Inspired by \cite{lee2013robust}, we use Expectation-Maximization (EM) algorithm to solve Eq.\eqref{eq::loss}. The latent weights $\mathcal{W}_\mathsf{K}$ and query poses $\mathcal{X}_\mathsf{K}$ are alternatively optimized until convergence (the change of $\mathcal{W}_\mathsf{K}$ is smaller than a threshold).

\paragraph{Expectation Step}
We aim to find the best $\mathcal{W}_\mathsf{K}$ while fixing $\mathcal{X}_\mathsf{K}$, the Expectation step becomes,
\begin{equation}\label{eq::ES}
    \argmin_{\mathcal{W}_\mathsf{K}}\sum_{k}w_k\left (\left \| \mathbf{P}_k - \mathbf{X}_k \right \|^2 + \lambda_1 \pi\left ( \mathbf{X}_k, \mathcal{F}_k\right )  \right ) - \underbrace{U^2\left (\log w_k - w_k\right)}_\text{Regularization} ,
\end{equation}
where the regularization term is added to avoid a  trivial solution of $\mathcal{W}_\mathsf{K} = \mathbf{0}$ and $U$ is a constant.

Eq.\eqref{eq::ES} is convex, and the minimum can be found by differentiating it with respect to $w_k$ and setting the gradient to zero. The updating of $w_k$ at the $t$-th iteration using query pose $\mathbf{X}_k^{t}$ is given by,
\begin{equation}
    w_k^{t+1} = \frac{U^2}{U^2+\left \| \mathbf{P}_k - \mathbf{X}_k^{t} \right \|^2 + \lambda_1 \pi\left ( \mathbf{X}_k^{t}, \mathcal{F}_k\right )}.
\end{equation}

\paragraph{Maximization Step}
We aim to find the best $\mathcal{X}_\mathsf{K}$ while fixing $\mathcal{W}_\mathsf{K}$, the updating of $\mathcal{X}_\mathsf{K}$ at the $t$-th iteration using weights $\mathcal{W}_\mathsf{K}^{t}$ is given by,
\begin{align}\label{eq::Ms}
    \mathcal{X}_\mathsf{K}^{t+1} = \argmin_{\mathcal{X}_\mathsf{K}}\sum_{k}&w_k^{t}\left (\left \| \mathbf{P}_k - \mathbf{X}_k \right \|^2 + \lambda_1 \pi\left ( \mathbf{X}_k, \mathcal{F}_k\right )  \right ) + \notag\\ &\left \|h\left (\mathbf{X}_{k-1}, \mathbf{X}_k  \right ) - \mathbf{Z}_{k-1,k}  \right \|^2  + \lambda_2\rho \left ( \mathbf{X}_{k-1}, \mathbf{X}_k, \mathcal{Q}_{k-1,k} \right ).
\end{align}

We solve \eqref{eq::Ms} using ceres \cite{Agarwal_Ceres_Solver_2022}.
\paragraph{Initialization} To initialize the iterative E-M updating process, we select a good subset of poses from $\mathcal{P}_K$ as seeds and initialize other query poses using their most recent seeds and relative SLAM poses, resulting in $ \mathcal{X}_\mathsf{K}^{1}$. A query pose with the number of 2D-3D matches ($\mathcal{F}_k$) larger than a pre-defined threshold is deemed as a seed.

\subsection{Experimental Result}
\label{sec:pgo_res}
Based on the global localization results of Sec. \ref{sec:cons_res}, we run the pose refinement step. Our refinement objective function has four terms (Eq.~\eqref{eq::loss}) and we conduct following ablation studies to validate the effectiveness of each term. Specifically,
\begin{enumerate}
    \item Pose Graph Optimization (PGO). By removing global 2D-3D reprojection term $\pi\left ( \mathbf{X}_k, \mathcal{F}_k\right )$ and 2D-2D Sampson term $\rho \left ( \mathbf{X}_{k-1}, \mathbf{X}_k, \mathcal{Q}_{k-1,k} \right )$, the objective function becomes PGO.  
    \item PGO\_2D2D. By removing global 2D-3D reprojection term $\pi\left ( \mathbf{X}_k, \mathcal{F}_k\right )$, the objective function becomes PGO with 2D-2D sampson term.
    \item PGO\_2D3D. By removing 2D-2D Sampson term $\rho \left ( \mathbf{X}_{k-1}, \mathbf{X}_k, \mathcal{Q}_{k-1,k} \right )$, the objective function becomes PGO with 2D-3D reprojection term.
    \item PGBA. We keep all four terms and denote the objective function as PGBA (PGO with Bundle Adjustment). 
\end{enumerate}
The results of the above four methods are given in Table \ref{tbl:pgo1} and \ref{tbl:pgo2}. 
For the oldtown dataset, both PGO\_2D2D and PGO\_2D3D outperforms PGO, showing the effectiveness of adding global 2D-3D reprojection and 2D-2D Sampson terms. The best performance is obtained by combining the two terms, resulting in PGBA.
For the cityloop dataset, all methods have similar performance probably because this dataset is saturated.

\begin{figure}[tbp]
       \centering
		\begin{subfigure}{0.8\textwidth}
			\centering
			\includegraphics[width=\textwidth]{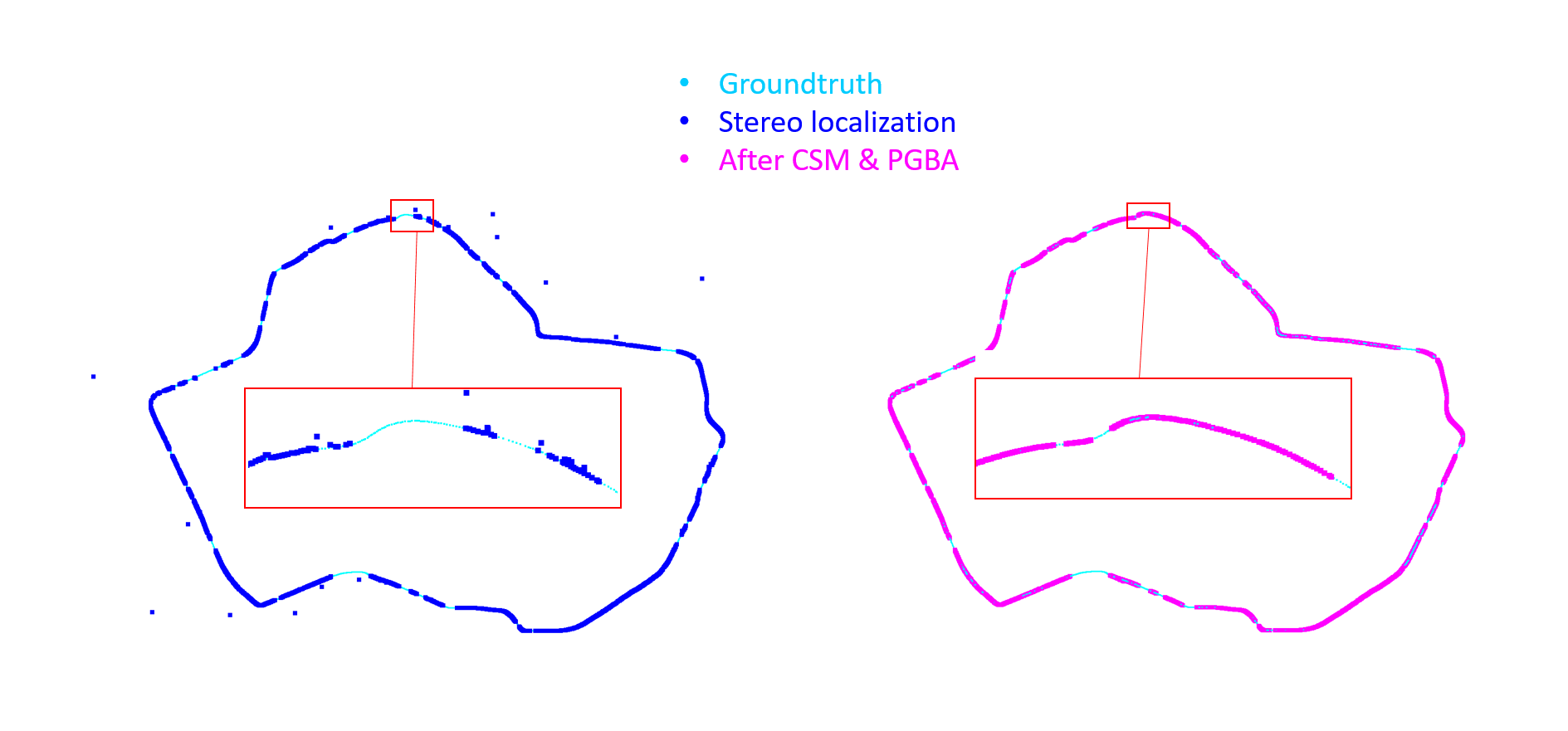}
			\caption{Oldtown}
		\end{subfigure}
		\begin{subfigure}{0.7\textwidth}
			\centering
			\includegraphics[width=\textwidth]{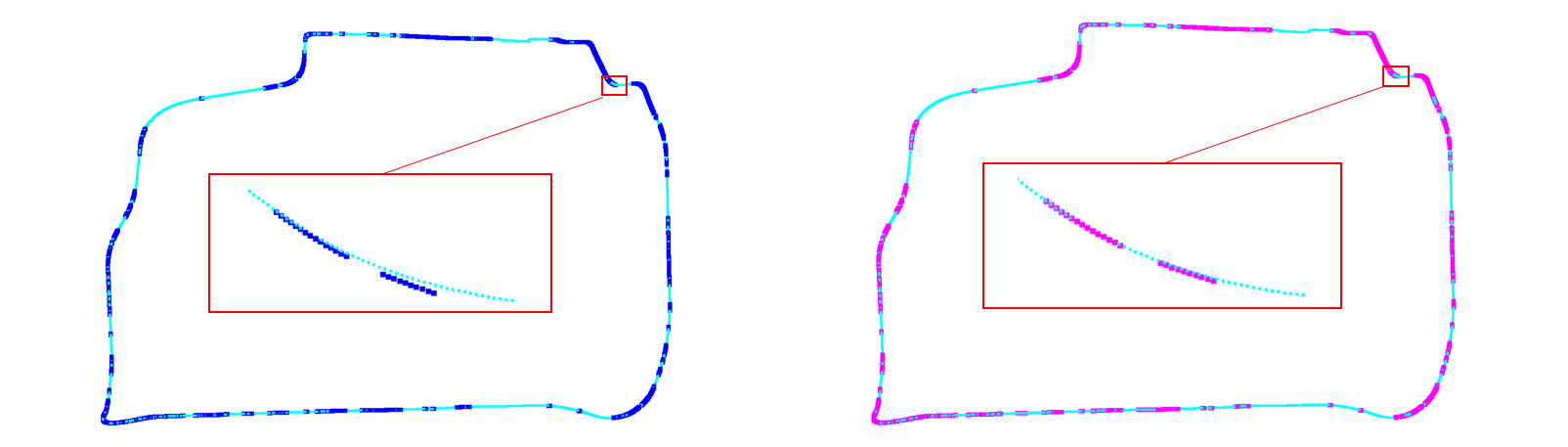}
			\caption{Cityloop}
		\end{subfigure}
		\caption{\it Query positions before and after refinement. Outlier positions are refined to correct ones, resulting in smooth trajectories.}
		\label{fig::refinement}
	\end{figure}

\begin{table}[t]
    \caption{\it Result of pose refinement on the oldtown datasets. Method PGBA achieves the best performance. }
    \vspace*{2mm}
    \label{tbl:pgo1}
    \centering
    \resizebox{0.9\textwidth}{!}{%
    \begin{tabularx}{\textwidth}{c *{5}{>{\centering\arraybackslash}X}} 
{Trans. Err.} & {Baseline} & {PGO} & PGO\_2D2D & PGO\_2D3D & PGBA \\ \hline
0.05m                & 49.80\%           & 55.10\%          & 56.20\%                                                          & 56.80\%          & \textbf{56.90\%}                                                  \\
0.1m                 & 78.40\%           & 83.80\%          & 84.30\%                                                          & 85.20\%          & \textbf{85.30\%}                                                  \\
0.2m                 & 93.30\%           & 96.80\%          & 96.80\%                                                          & 97.20\%          & \textbf{97.30\%}                                                  \\
0.5m                 & 96.50\%           & \textbf{98.50\%} & 98.40\%                                                          & 98.40\%          & 98.40\%                                                           \\
1.0m                 & 97.80\%           & \textbf{99.00\%} & \textbf{99.00\%}                                                 & \textbf{99.00\%} & \textbf{99.00\%}                                                  \\
3.0m                 & 98.30\%           & \textbf{99.30\%} & \textbf{99.30\%}                                                 & \textbf{99.30\%} & \textbf{99.30\%}                                                  \\ \hline

    \end{tabularx}
    }

\end{table}
\begin{table}[tbp]
    \caption{\it {Result of pose refinement on the cityloop dataset.} Method PGBA achieves the best performance.}
    \vspace*{2mm}
    \label{tbl:pgo2}
    \centering
    \resizebox{0.9\textwidth}{!}{%
    \begin{tabularx}{\textwidth}{c *{5}{>{\centering\arraybackslash}X}} 
{Trans. Err.} & {Baseline} & {PGO} & PGO\_2D2D & PGO\_2D3D & PGBA \\ \hline
0.05m                & 79.20\%                               & 81.00\%                          & 81.00\%                                                                              & 81.30\%                           & \textbf{81.40\%}                                                                      \\
0.1m                 & 92.30\%                               & 93.00\%                          & 93.00\%                                                                              & \textbf{93.10\%}                  & 93.00\%                                                                               \\
0.2m                 & 97.70\%                               & 98.20\%                          & 98.20\%                                                                              & \textbf{98.60\%}                  & \textbf{98.60\%}                                                                      \\
0.5m                 & \textbf{99.70\%}                      & \textbf{99.70\%}                 & \textbf{99.70\%}                                                                     & \textbf{99.70\%}                  & \textbf{99.70\%}                                                                      \\
1.0m                 & \textbf{100.00\%}                     & \textbf{100.00\%}                & \textbf{100.00\%}                                                                    & \textbf{100.00\%}                 & \textbf{100.00\%}                                                                     \\
3.0m                 & \textbf{100.00\%}                     & \textbf{100.00\%}                & \textbf{100.00\%}                                                                    & \textbf{100.00\%}                 & \textbf{100.00\%}                                                                     \\ \hline

    \end{tabularx}
    }

\end{table}
We show estimated query positions before and after consensus set maximization and pose refinement steps in Fig.~\ref{fig::refinement}. It clearly shows that wrong estimated query positions are refined to correct positions, resulting in a smooth trajectory after refinement.

Our pose refinement step is implemented using C++, and the processing time is given in Table~\ref{tab:pgo_runtime}. For time-critical applications, methods PGO and PGO\_2D3D are good candidates. 

\begin{table}[tbp]
\centering
\caption{\it {Running time of pose refinement}. We have 3296 and 10224  query frames for the oldtown and cityloop dataset, respectively. Since initial poses of the cityloop dataset are more accurate than those of oldtown dataset, fewer EM iterations are performed for cityloop, resulting in its smaller running time. }
\label{tab:pgo_runtime}
    \vspace*{2mm}

\resizebox{0.9\textwidth}{!}{%
\begin{tabularx}{1.0\textwidth}{ll *{4}{>{\centering\arraybackslash}X}} 
{Dataset }  && {PGO} & PGO\_2D2D & PGO\_2D3D & PGBA \\ \toprule
Oldtown &&  26.7s& 2486.7s &102.0s& 2245.0s\\ \hline
Cityloop  &&  21.0s & 2263.8s &93.9s& 2677.8s \\ \bottomrule
\end{tabularx}
}
\end{table}


\section{Pose Polishing}
Note that the global pose $\mathbf{P}_k$ used in the pose refinement (Sec.\ref{sec::pose_refinement}) step is from one of the multi-sessions localization results. Using optimized query poses $\mathcal{X}_\mathsf{K}$ from Sec.\ref{sec::pose_refinement}, we can unify information from multi-sessions to obtain a better $\mathbf{P}_{k,i}$, followed by another round of consensus set maximization and pose refinement as described in Sec.\ref{sec::Multi-sessionCSM} and Sec.\ref{sec::pose_refinement}. This final pose polishing step is performed (optionally)  only once.

As discussed in Sec.\ref{sec::Multi-sessionCSM}, simply combining all 2D-3D matches from multi-session maps would produce an averaged (or one dominant) pose over multi-session poses. Using the accurate $\mathcal{X}_\mathsf{K}$ as an anchor, we first compute reprojection errors of all 2D-3D matches from multi-session maps and then prune out outlier 2D-3D matches. Remaining 2D-3D matches are used to compute the final global pose $\mathbf{P}_{k_i}$ for the final round of pose fusion and refinement. This guided pose polish step could further improve the final pose accuracy for better application.

\subsection{Experimental Result}
Based on estimated poses from Sec. \ref{sec:pgo_res}, we test the pose polish module.
We use refined poses $\mathcal{X}_\mathsf{K}$ to filter 2D-3D matches and recompute the pose for each session. A second round of consensus set maximization and pose refinement is performed on these poses. The results are given in Table \ref{tbl:polish}.
For both datasets, polishing helps to further improve pose accuracies, as the Recall@0.05m increases. 
\begin{table}[tbp]
    \caption{\it Results of pose polish on the oldtown and cityloop datasets. With the second round of consensus set maximization and pose refinement, we get more accurate poses.}
    \vspace*{2mm}
    \label{tbl:polish}
    \centering
    \resizebox{0.9\textwidth}{!}{%
    \begin{tabularx}{\textwidth}{c *{5}{>{\centering\arraybackslash}X}} 
\multicolumn{1}{l}{} & \multicolumn{2}{c}{{Oldtown}} & \multicolumn{2}{c}{{Cityloop}} \\ \cline{2-5} 
{Trans. Err.} & {Baseline}  & {Polish} & {Baseline}  & {Polish}  \\ \hline
0.05m                & 56.90\%            &  \textbf{59.70}\%         & 81.40\%            &  \textbf{82.10}\%          \\
0.1m                 & 85.30\%            & 85.20\%         & 93.00\%            &  \textbf{93.30}\%          \\
0.2m                 & 97.30\%            & 97.20\%         & 98.60\%            & 98.60\%          \\
0.5m                 & 98.40\%            &  \textbf{98.50}\%         & 99.70\%            & 99.70\%          \\
1.0m                 & 99.00\%            & 99.00\%         & 100.00\%           & 100.00\%         \\
3.0m                 & 99.30\%            & 99.30\%         & 100.00\%           & 100.00\%         \\ \hline
    \end{tabularx}
    }

\end{table}

\section{Conclusions}
We have proposed a method, named \textit{CyberLoc}, for robust and accurate visual localization under challenging conditions. The key idea is to build a robust map for each reference sequence, find the best global camera pose with respect to multi-session maps, and combine global localization and local SLAM to refine camera poses. The proposed robust mapping and localization, consensus set maximization and pose refinement facilitates the success of our method, to be used in scenes with illumination and environmental changes. Extensive experiments on 4seasons datasets demonstrate the high accuracy and robustness of our method.

\clearpage

\bibliographystyle{splncs}
\bibliography{egbib}

\begin{thebibliography}{10}

\bibitem{wenzel20204seasons}
Wenzel, P., Wang, R., Yang, N., Cheng, Q., Khan, Q., von Stumberg, L., Zeller,
  N., Cremers, D.:
\newblock 4seasons: A cross-season dataset for multi-weather slam in autonomous
  driving.
\newblock In: DAGM German Conference on Pattern Recognition, Springer (2020)
  404--417

\bibitem{strisciuglio2018trimbot2020}
Strisciuglio, N., Tylecek, R., Blaich, M., Petkov, N., Biber, P., Hemming, J.,
  van Henten, E., Sattler, T., Pollefeys, M., Gevers, T.,  et~al.:
\newblock Trimbot2020: An outdoor robot for automatic gardening.
\newblock In: ISR 2018; 50th International Symposium on Robotics, VDE (2018)
  1--6

\bibitem{sarlin2022lamar}
Sarlin, P.E., Dusmanu, M., Sch{\"o}nberger, J.L., Speciale, P., Gruber, L.,
  Larsson, V., Miksik, O., Pollefeys, M.:
\newblock Lamar: Benchmarking localization and mapping for augmented reality.
\newblock arXiv preprint arXiv:2210.10770 (2022)

\bibitem{wang2021low}
Wang, Y., Wan, R., Yang, W., Li, H., Chau, L.P., Kot, A.C.:
\newblock Low-light image enhancement with normalizing flow.
\newblock arXiv preprint arXiv:2109.05923 (2021)

\bibitem{liu2017robust}
Liu, L., Li, H., Dai, Y., Pan, Q.:
\newblock Robust and efficient relative pose with a multi-camera system for
  autonomous driving in highly dynamic environments.
\newblock IEEE Transactions on Intelligent Transportation Systems
  \textbf{19}(8) (2017)  2432--2444

\bibitem{saputra2018visual}
Saputra, M.R.U., Markham, A., Trigoni, N.:
\newblock Visual slam and structure from motion in dynamic environments: A
  survey.
\newblock ACM Computing Surveys (CSUR) \textbf{51}(2) (2018)  1--36

\bibitem{xie2021segformer}
Xie, E., Wang, W., Yu, Z., Anandkumar, A., Alvarez, J.M., Luo, P.:
\newblock Segformer: Simple and efficient design for semantic segmentation with
  transformers.
\newblock Advances in Neural Information Processing Systems \textbf{34} (2021)
  12077--12090

\bibitem{arandjelovic2016netvlad}
Arandjelovic, R., Gronat, P., Torii, A., Pajdla, T., Sivic, J.:
\newblock Netvlad: Cnn architecture for weakly supervised place recognition.
\newblock In: Proceedings of the IEEE conference on computer vision and pattern
  recognition. (2016)  5297--5307

\bibitem{liu2019stochastic}
Liu, L., Li, H., Dai, Y.:
\newblock Stochastic attraction-repulsion embedding for large scale image
  localization.
\newblock In: Proceedings of the IEEE/CVF International Conference on Computer
  Vision. (2019)  2570--2579

\bibitem{ge2020self}
Ge, Y., Wang, H., Zhu, F., Zhao, R., Li, H.:
\newblock Self-supervising fine-grained region similarities for large-scale
  image localization.
\newblock In: European conference on computer vision, Springer (2020)  369--386

\bibitem{detone2018superpoint}
DeTone, D., Malisiewicz, T., Rabinovich, A.:
\newblock Superpoint: Self-supervised interest point detection and description.
\newblock In: Proceedings of the IEEE conference on computer vision and pattern
  recognition workshops. (2018)  224--236

\bibitem{schonberger2016structure}
Schonberger, J.L., Frahm, J.M.:
\newblock Structure-from-motion revisited.
\newblock In: Proceedings of the IEEE conference on computer vision and pattern
  recognition. (2016)  4104--4113

\bibitem{chen2021learning}
Chen, H., Luo, Z., Zhang, J., Zhou, L., Bai, X., Hu, Z., Tai, C.L., Quan, L.:
\newblock Learning to match features with seeded graph matching network.
\newblock In: Proceedings of the IEEE/CVF International Conference on Computer
  Vision. (2021)  6301--6310

\bibitem{shi2022clustergnn}
Shi, Y., Cai, J.X., Shavit, Y., Mu, T.J., Feng, W., Zhang, K.:
\newblock Clustergnn: Cluster-based coarse-to-fine graph neural network for
  efficient feature matching.
\newblock arXiv preprint arXiv:2204.11700 (2022)

\bibitem{suwanwimolkul2022efficient}
Suwanwimolkul, S., Komorita, S.:
\newblock Efficient linear attention for fast and accurate keypoint matching.
\newblock arXiv preprint arXiv:2204.07731 (2022)

\bibitem{thomee2016yfcc100m}
Thomee, B., Shamma, D.A., Friedland, G., Elizalde, B., Ni, K., Poland, D.,
  Borth, D., Li, L.J.:
\newblock Yfcc100m: The new data in multimedia research.
\newblock Communications of the ACM \textbf{59}(2) (2016)  64--73

\bibitem{bian2019evaluation}
Bian, J.W., Wu, Y.H., Zhao, J., Liu, Y., Zhang, L., Cheng, M.M., Reid, I.:
\newblock An evaluation of feature matchers for fundamental matrix estimation.
\newblock arXiv preprint arXiv:1908.09474 (2019)

\bibitem{FGCNet}
Liu, L., Liyuan, P., Wei, L., Qichao, X., Yuxiang, W., Jiangwei, L.:
\newblock Fgcnet: Fast graph convolution for matching features.
\newblock ISMAR (2022)

\bibitem{peng2021megloc}
Peng, S., He, Z., Zhang, H., Yan, R., Wang, C., Zhu, Q., Liu, X.:
\newblock Megloc: A robust and accurate visual localization pipeline.
\newblock arXiv preprint arXiv:2111.13063 (2021)

\bibitem{zou2012coslam}
Zou, D., Tan, P.:
\newblock Coslam: Collaborative visual slam in dynamic environments.
\newblock IEEE transactions on pattern analysis and machine intelligence
  \textbf{35}(2) (2012)  354--366

\bibitem{pire2017s}
Pire, T., Fischer, T., Castro, G., De~Crist{\'o}foris, P., Civera, J., Berlles,
  J.J.:
\newblock S-ptam: Stereo parallel tracking and mapping.
\newblock Robotics and Autonomous Systems \textbf{93} (2017)  27--42

\bibitem{tang2022quadtree}
Tang, S., Zhang, J., Zhu, S., Tan, P.:
\newblock Quadtree attention for vision transformers.
\newblock arXiv preprint arXiv:2201.02767 (2022)

\bibitem{zhou2021patch2pix}
Zhou, Q., Sattler, T., Leal-Taixe, L.:
\newblock Patch2pix: Epipolar-guided pixel-level correspondences.
\newblock In: Proceedings of the IEEE/CVF conference on computer vision and
  pattern recognition. (2021)  4669--4678

\bibitem{sarlin2020superglue}
Sarlin, P.E., DeTone, D., Malisiewicz, T., Rabinovich, A.:
\newblock Superglue: Learning feature matching with graph neural networks.
\newblock In: Proceedings of the IEEE/CVF conference on computer vision and
  pattern recognition. (2020)  4938--4947

\bibitem{cheng2020hierarchical}
Cheng, X., Zhong, Y., Harandi, M., Dai, Y., Chang, X., Li, H., Drummond, T.,
  Ge, Z.:
\newblock Hierarchical neural architecture search for deep stereo matching.
\newblock Advances in Neural Information Processing Systems \textbf{33} (2020)

\bibitem{pless2003using}
Pless, R.:
\newblock Using many cameras as one.
\newblock In: 2003 IEEE Computer Society Conference on Computer Vision and
  Pattern Recognition, 2003. Proceedings. Volume~2., IEEE (2003)  II--587

\bibitem{labbe2021multi}
Labb{\'e}, M., Michaud, F.:
\newblock Multi-session visual slam for illumination invariant localization in
  indoor environments.
\newblock arXiv preprint arXiv:2103.03827 (2021)

\bibitem{hartley2003multiple}
Hartley, R., Zisserman, A.:
\newblock Multiple view geometry in computer vision.
\newblock Cambridge university press (2003)

\bibitem{li2009consensus}
Li, H.:
\newblock Consensus set maximization with guaranteed global optimality for
  robust geometry estimation.
\newblock In: 2009 IEEE 12th International Conference on Computer Vision, IEEE
  (2009)  1074--1080

\bibitem{lee2013robust}
Lee, G.H., Fraundorfer, F., Pollefeys, M.:
\newblock Robust pose-graph loop-closures with expectation-maximization.
\newblock In: 2013 IEEE/RSJ International Conference on Intelligent Robots and
  Systems, IEEE (2013)  556--563

\bibitem{Agarwal_Ceres_Solver_2022}
Agarwal, S., Mierle, K., Team, T.C.S.:
\newblock {Ceres Solver} (3 2022)

\end{thebibliography}
\end{document}